\pdfoutput=1

\documentclass[11pt]{article}

\usepackage{ACL2023}

\usepackage{times}
\usepackage{latexsym}

\usepackage{tabularx}
\usepackage[normalem]{ulem}

\usepackage[T1]{fontenc}

\usepackage[utf8]{inputenc}

\usepackage{microtype}

\usepackage{inconsolata}

\usepackage{float}
\usepackage{chngcntr}
\usepackage[super]{nth}
\usepackage{nicefrac}
\usepackage{microtype}

\usepackage{amsmath,amssymb,amsfonts}
\usepackage{graphicx}
\usepackage{textcomp}
\usepackage{xcolor}
\usepackage{lipsum}

\usepackage{microtype}
\usepackage{booktabs} 
\usepackage{vcell}
\usepackage{amsmath}
\usepackage{amssymb}

\usepackage{graphicx}
\usepackage{subfigure}

\usepackage{multirow}
\usepackage{booktabs}
\usepackage{colortbl}
\usepackage{graphicx}
\usepackage{subfigure}
\usepackage{lipsum}
\usepackage{nicefrac}
\usepackage{float}
\usepackage{chngcntr}
\usepackage[super]{nth}
\usepackage{nicefrac}
\usepackage{microtype}

\newcommand{\comment}[1]{}

\usepackage{amsmath}

\newcommand{\ie}{\textit{i.e.}}

\usepackage{hyperref}

\usepackage[ruled,vlined,linesnumbered]{algorithm2e}

\usepackage{hyperref}

\title{Guiding Text-to-Text Privatization by Syntax}

\author{Stefan Arnold \and Dilara Yesilba \and Sven Weinzierl \\ Friedrich-Alexander-Universität Erlangen-Nürnberg \\ Lange Gasse 20, 90403 Nürnberg, Germany \\ \texttt{(stefan.st.arnold, dilara.yesilbas, sven.weinzierl)@fau.de}}

\begin{document}

\maketitle

\begin{abstract}

    \textit{Metric Differential Privacy} is a generalization of differential privacy tailored to address the unique challenges of text-to-text privatization. By adding noise to the representation of words in the geometric space of embeddings, words are replaced with words located in the proximity of the noisy representation. Since embeddings are trained based on word co-occurrences, this mechanism ensures that substitutions stem from a common semantic context. Without considering the grammatical category of words, however, this mechanism cannot guarantee that substitutions play similar syntactic roles. We analyze the capability of text-to-text privatization to preserve the grammatical category of words after substitution and find that surrogate texts consist almost exclusively of nouns. Lacking the capability to produce surrogate texts that correlate with the structure of the sensitive texts, we encompass our analysis by transforming the privatization step into a candidate selection problem in which substitutions are directed to words with matching grammatical properties. We demonstrate a substantial improvement in the performance of downstream tasks by up to $4.66\%$ while retaining comparative privacy guarantees.
\end{abstract}

\section{Introduction}
\label{section:1}

From compliance with stringent data protection regulations to building trust, privacy emerged as a formidable challenge to applications that build on user-generated data, and consensus exists regarding the need to safeguard user privacy.

In the context of text analysis, privacy is typically protected by sanitizing personally identifiable information from the text via ad-hoc filtering or anonymization. The literature is replete with naïve approaches that either redact words from the text or insert distractive words into the text. Using generalization and suppression on quasi-identifiers, an intuitive way of expressing privacy is presented by $k$-anonymity \citep{sweeney2002k} and its notable adaptations for text data \citep{jiang2009t, sanchez2016c}.

However, these approaches are fundamentally flawed. Incapable of anticipating an adversary's side knowledge, most anonymization schemes are vulnerable to re-identification and thus provably non-private. As text conveys seemingly innocuous information, researchers demonstrated that this information can be leveraged to identify authorship \citep{song2019auditing} or disclose identifiable information \citep{carlini2020extracting, pan2020privacy, song2020information, thomas2020investigating}. \citet{carlini2020extracting}, for instance, recovered verbatim text from the training corpus using black-box querying to a language model. 

Building upon noise calibration, \textit{Differential Privacy} (DP) \citep{dwork2006calibrating} attracted considerable attention for their robust notion of privacy. For text analysis, DP is applied to the vector-valued representation of text data \citep{coavoux2018privacy,weggenmann2018syntf,vu2019dpugc}.

We focus on \textit{Metric Differential Privacy} \citep{chatzikokolakis2013broadening}, in which data is processed independently, similar to the setting of randomized response \citep{kasiviswanathan2011can}. To avoid the curse of dimensionality of randomized response, noise is scaled by a general distance metric. For text-to-text privatization, \citet{feyisetan2020privacy} adopted a distance metric so that words that are close (\textit{i.e.} more similar) to a word are assigned with a higher substitution probability than those that are more distant (\textit{i.e.} less similar). This requires that the text is mapped onto a continuous embedding space \citep{mikolov2013efficient, pennington2014glove, bojanowski2017enriching}. Proceeding from the embedding, each word in the text is privatized by a three-step protocol: (1) retrieving the vector representation of the word, (2) perturbing the vector representation of the word with noise sampled from a multivariate distribution, and (3) projecting the noisy representation of the word back to the discrete vocabulary space. As the noisy representations are unlikely to exactly represent words in the embedding space, a nearest neighbor approximation is returned. 

Since text-to-text privatization operates directly on embeddings and words in the embedding space are mapped based on co-occurrences, words tend to be substituted by words that stem from a common semantic context. However, there is no guarantee that words are substituted by words that serve similar roles within the grammatical structure of a text. Motivated by the example of sentiment analysis, in which sentiment is typically expressed by adjectives and forms of adjectives \citep{benamara2007sentiment}, we hypothesize that substitutions strictly based on co-occurrences may degrade downstream performance. This hypothesis is in line with linguists finding repeated evidence for the relevance of grammatical properties for language understanding \citep{myhill2012re}. 

We summarize our contributions as follows:

\begin{enumerate}

\item[$\bullet$] We investigate text-to-text privatization via metric differential privacy in terms of its capability to preserve the grammatical properties of words after substitution. We find that privatization produces texts that consist to a large extent of incoherent nouns.

\item[$\bullet$] We incorporate grammatical categories into the privatization step in the form of a constraint to the candidate selection. We demonstrate that broadening the candidate pool  to $k>1$ (instead of $k=1$) and selecting a substitution with matching grammatical properties amplifies the performance in downstream tasks while maintaining an equivalent level of privacy.


\end{enumerate}

\section{Preliminaries}
\label{section:2}

\subsection{Differential Privacy}

\textit{Differential Privacy} (DP) \citep{dwork2006calibrating} emerged as a robust notion for privacy applied in privacy-preserving data mining and machine learning. Due to its composability and robustness to post-processing regardless of an adversary’s side knowledge, it formalizes privacy without the critical pitfalls of previous anonymization schemes. To ensure a consistent understanding of the algorithmic foundation of differential privacy, we present a brief taxonomy and a formal definition of the variants used for text analysis. 

Formally, a randomized mechanism $\mathcal{M}: \mathcal{D} \rightarrow \mathcal{R}$ with domain $\mathcal{D}$ and range $\mathcal{R}$ satisfies $\varepsilon$-indistinguishability if any two adjacent inputs $d,d’ \in \mathcal{D}$ and for any subset of outputs $S \subseteq \mathcal{R}$ it holds that:
\begin{equation}
\label{equation:dp1}
\frac{\mathbb{P}[\mathcal{M}(d) \in S]}{\mathbb{P}[\mathcal{M}(d') \in S]} \leq e^{\varepsilon}.
\end{equation}

At a high level, a randomized mechanism is differentially-private if the output distributions from two adjacent datasets are (near) indistinguishable, where any two datasets are considered adjacent that differ in at most one record. An adversary seeing the output can therefore not discriminate if a particular observation was used. This notion of indistinguishability is controlled by the parameter $\varepsilon$ acting as a privacy budget. It defines the strength of the privacy guarantee (with $\varepsilon \rightarrow 0$ representing strict privacy and $\varepsilon \rightarrow \infty$ representing the lack of privacy). To enhance the accounting of the privacy budget, several relaxations exist \citep{dwork2006our,mironov2017renyi,dong2019gaussian}.

Depending on the setting, DP can be categorized into \textit{global} DP \citep{dwork2006calibrating} and \textit{local} DP \citep{kasiviswanathan2011can}. 

Global DP addresses the setting in which privacy is defined with respect to aggregate statistics. It assumes a trusted curator who can collect and access raw user data. The randomized mechanism is applied to the collected dataset to produce differentially private output for downstream use. With noise drawn from a predetermined distribution, the design of the randomized mechanism builds upon an additive noise mechanism. Commonly used distributions for adding noise include Laplace and Gaussian distribution \citep{dwork2014algorithmic}. The noise is further calibrated according to the function’s sensitivity and the privacy budget. This technique is useful for controlling the disclosure of private information of records processed with real-valued and vector-valued functions. 

Local DP addresses the setting in which privacy is defined with respect to individual records. In contrast to global DP, local DP does not rely on a trusted curator. Instead of a trusted curator that applies the randomized mechanism, the randomized mechanism is applied to all records independently to provide plausible deniability \citep{bindschaedler2017plausible}. The randomized mechanism to achieve local DP is typically \textit{Randomized Response} (RR) \citep{warner1965randomized}, which protects private information by answering a plausible response to the sensitive query.

Since we aim for text-to-text privatization, formulating DP in the local setting through RR appears to be a natural solution. However, the strong privacy guarantees constituted by RR impose requirements that render it impractical for text. That is, RR requires that a sentence $s$ must have a non-negligible probability of being transformed into any other sentence $s^{'}$ regardless of how unrelated $s$ and $s^{'}$ are. This indistinguishability constraint makes it virtually impossible to enforce that the semantics of a sentence $s$ are approximately captured by a privatized sentence $s^{'}$. Since the vocabulary size can grow exponentially large in length $|s|$, the number of sentences semantically related to $s$ becomes vanishingly small probability under RR \citep{feyisetan2020privacy}.

\subsection{Metric Differential Privacy}

\textit{Metric Differential Privacy} \citep{chatzikokolakis2013broadening} is a generalization of differential privacy that originated in the context of location-based privacy, where locations close to a user are assigned with a high probability, while distant locations are given negligible probability. By using word embeddings as a corollary to geo-location coordinates, metric differential privacy was adopted from location analysis to textual analysis by \citet{feyisetan2020privacy}. 

We follow the formulation of \citet{xu2021utilitarian} for metric differential privacy in the context of textual analysis. Equipped with a discrete vocabulary set $\mathcal{W}$, an embedding function $\phi : \mathcal{W} \rightarrow \mathbb{R}$, where $\mathbb{R}$ represents a high-dimensional embedding space, and a distance function $d: \mathbb{R} \times \mathbb{R} \rightarrow [0,\infty)$ satisfying the axioms of a metric (\ie, identity of indiscernibles, symmetry, and triangle inequality), metric differential privacy is defined in terms of the distinguishability level between pairs of words. A randomized mechanism $\mathcal{M}:\mathcal{W} \rightarrow \mathcal{W}$ satisfies metric differential privacy with respect to the distance metric $d(\cdot)$ if for any $w,w^{'},\hat{w} \in \mathcal{W}$ the output distributions of $\mathcal{M}(w)$ and $\mathcal{M}(w^{'})$ are bounded by Equation \ref{equation:dp2} for any privacy budget $\varepsilon > 0$:

\begin{equation}
\label{equation:dp2}
 \frac{\mathbb{P} [\mathcal{M}(w) = \hat{w}]}{\mathbb{P} [\mathcal{M}(w^{'}) = \hat{w}]} \leq e^{\varepsilon d\{\phi(w),\phi(w^{'})\}}.
\end{equation}

This probabilistic guarantee ensures that the log-likelihood ratio of observing any word $\hat{w}$ given two words $w$ and $w’$ is bounded by $\varepsilon d\{\phi(w),\phi(w’)\}$ and provides plausible deniability \citep{bindschaedler2017plausible} with respect to all $w \in \mathcal{W}$. We refer to \citet{feyisetan2020privacy} for a complete proof of privacy. For $\mathcal{M}$ to provide plausible deniability, additive noise is in practice sampled from a multivariate distribution such as the \textit{multivariate Laplace distribution} \citep{feyisetan2020privacy} or \textit{truncated Gumbel distribution} \citep{carvalho2021tem}.

We recall that differential privacy requires adjacent datasets that differ in at most one record. Since the distance $d(\cdot)$ captures the notion of closeness between datasets, metric differential privacy instantiates differential privacy when Hamming distance is used, \ie, if $\forall x,x': d\{\phi(w),\phi(w^{'})\} = 1$. Depending on the distance function $d(\cdot)$, metric differential privacy is therefore generally less restrictive than differential privacy. Intuitively, words that are distant in metric space are easier to distinguish compared words that are in close proximity. Scaling the indistinguishability by a distance $d(\cdot)$ avoids the curse of dimensionality that arises from a large vocabulary $\mathcal{W}$ and allows the mechanism $\mathcal{M}$ to produce similar substitutions $\hat{w}$ for similar $w$ and $w^{'}$. However, this scaling complicates the interpretation of the privacy budget $\varepsilon$, as it changes depending on the metric employed.

\subsection{Related Work}
\label{section:3}

Grounded in metric differential privacy, text-to-text privatization implies that the indistinguishability of substitutions of any two words in the vocabulary is scaled by their distance. 

\citet{fernandes2018author} achieve this indistinguishability by generating a bag-of-words representation and applying the \textit{Earth Mover’s distance} to obtain privatized bags. 

In contrast to a bag-of-words representation, \citet{feyisetan2020privacy} formalized text-to-text privatization to operate on continuous word embeddings. Word embeddings capture the level of semantic similarity between words and have been popularized by efficient embedding mechanisms \citep{mikolov2013efficient, pennington2014glove}. This mechanism was termed \texttt{MADLIB}. 


The issue with this mechanism is that the magnitude of the noise is proportional to the dimensionality of the vector representation. This translates into adding the same amount of noise to any word in the embedding space, regardless of whether this word is located in a dense or sparse region. For words in densely populated areas, adding noise that is large in magnitude renders it difficult for the mechanism to select reasonable substitutions, as nearby relevant words cannot be distinguished from other nearby but irrelevant words. For words in sparsely populated areas, adding noise of small magnitude renders the mechanism susceptible to reconstruction, as the word closest to a noisy representation is likely to be the original word.

To tackle some of the severe shortcomings of \texttt{MADLIB}, a variety of distance metrics have been employed to scale the indistinguishability, including \textit{Hamming distance} \citep{carvalho2021brr}, \textit{Manhattan distance} \citep{fernandes2019generalised}, \textit{Euclidean distance} \citep{fernandes2019generalised,feyisetan2020privacy,carvalho2021tem, feyisetan2021private}, \textit{Mahalanobis distance} \citep{xu2020differentially} and \textit{Hyperbolic distance} \citep{feyisetan2019leveraging}. 

While related extensions have focused almost exclusively on geometric properties to enhance text-to-text privatization, we focus on linguistic properties. We extend \texttt{MADLIB} by a candidate selection that directs substitutions based on matching grammatical properties and demonstrate that multivariate perturbations supported by grammatical properties substantially improve the utility of the surrogate texts in downstream tasks.

\begin{figure}[!t]
    \centering
    \includegraphics[width=0.5\textwidth]{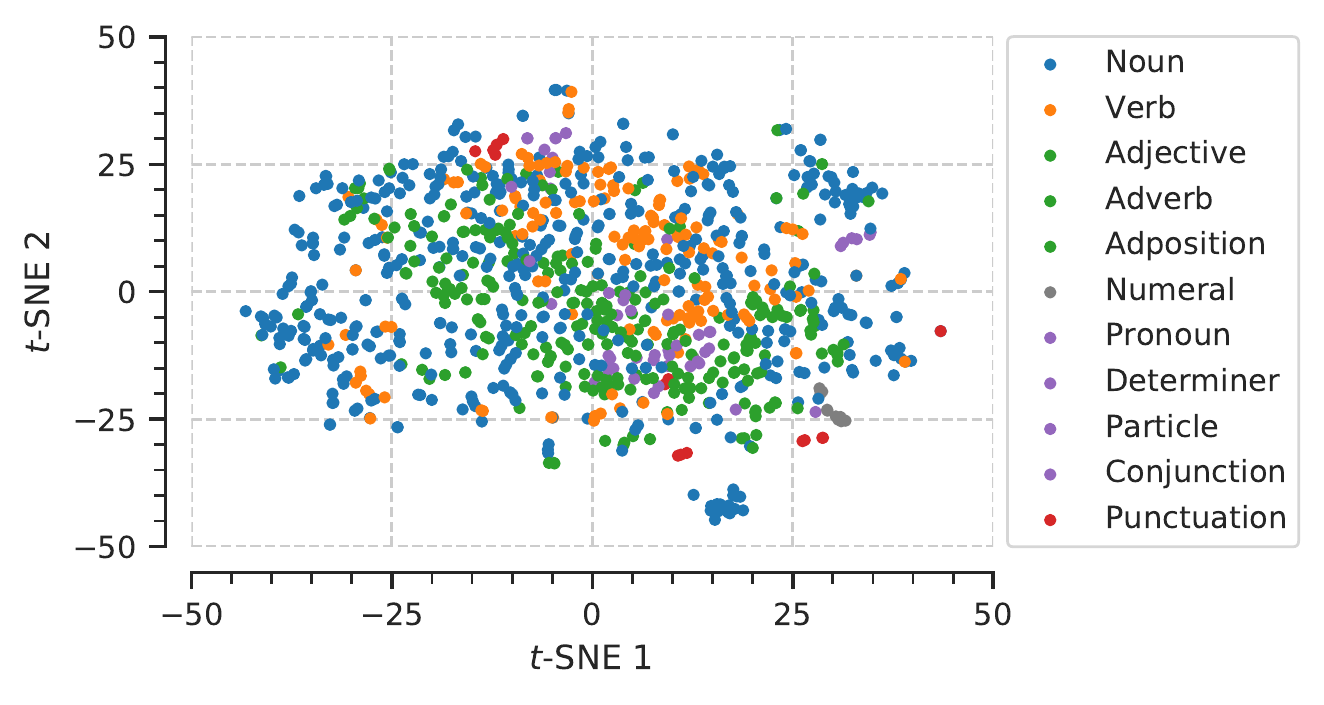}
    \caption{Embedding space of the $1,000$ most frequent words in $100$-dimensional \texttt{GloVe}, automatically encoded with their universal part-of-speech tags.}
    \label{fig:embedding}
\end{figure}


\section{Methodology}
\label{section:4}

Since text-to-text privatization operates directly on geometric space of embeddings, it is necessary to understand the structure of the embedding space. To get an understanding of the embedding space, we selected a subset of $1,000$ most frequent words from the $100$-dimensional \texttt{GloVe} embedding and manifolded them onto a two-dimensional representation. Enriched by grammatical properties derived from the universal part-of-speech tagset \citep{petrov2011universal}, we chart a $t$-distributed stochastic neighbor embedding \citep{van2008visualizing} in Figure \ref{fig:embedding}. 

We note that we derived each word's grammatical category without context, which may explain the general tendency towards \textit{nouns} (presumably misclassified \textit{verbs}). Regardless of potentially misclassified grammatical categories, we can draw the following conclusions: while \textit{nouns}, \textit{verbs}, and \textit{adjectives} are distributed throughout the embedding space, we find distinct subspaces for \textit{numerals} and \textit{punctuation}. This is because word embeddings are trained towards an objective that ensures that words occurring in a common context have similar embeddings, disregarding their syntactic roles within the structure of a text. Considering that text-to-text privatization typically selects the nearest approximate neighbor after the randomized mechanism is queried as substitution, we expect this mechanism to fall short in producing syntactically coherent texts.

We adopt the multivariate Laplace mechanisms of \texttt{MADLIB} \citep{feyisetan2020privacy}. Aimed at preserving the grammatical category of a word after its substitution, we incorporate a constraint into the candidate selection that directs the randomized mechanism towards words with a matching grammatical category. This constraint is incorporated as follows: we create a dictionary that serves as a lookup table for the grammatical category of each word in the vocabulary and generalize the randomized mechanism to return a flexible $k \gg 1$ (instead of $k=1$) approximate nearest neighbors. If available, a word is replaced by the nearest word (measured from the noisy representation) that matches its grammatical category. Otherwise, the protocol reduces to canonical \texttt{MADLIB}. The computational overhead of the candidate selection is $O(\log k)$. 

This modification introduces the size of the candidate pool $k$ as an additional hyperparameter. Intuitively, $k$ should be chosen based on the geometric properties of the embedding, \ie, $k$ should be large enough to contain at least one other word with a matching grammatical category. 

We investigate our modification to\texttt{MADLIB} in terms of its capability to preserve grammatical properties and its implications. For reasons of reproducibility, we base all experiments on the $100$-dimensional \texttt{GloVe} embedding. 


To keep the computational effort feasible, we formed a vocabulary that consists of $24,525$ words reflecting a natural distribution of grammatical categories: $26$ \textit{pronouns}, $5,000$ \textit{nouns}, $5,000$ \textit{verbs}, $5,000$ \textit{adjectives}, $4,341$ \textit{adverbs}, $92$ \textit{adpositions}, $5,000$ \textit{numerals}, $6$ \textit{conjunctions}, $2$ \textit{particles}, $39$ \textit{determiner}, and $19$ \textit{punctuations}.

Once we determined our sub-vocabulary, we calculated the necessary size of the candidate pool $k$. We counted the number of steps required from each word in our subset until a neighbor with a matching category was found. Averaging this count revealed that each word is linked to another word with a matching category within a neighborhood of $20$. We thus parameterized the candidate pool to a fixed $k=20$ across all experiments. 





\section{Experiments}

We conduct a series of experiments at a strategically chosen set of privacy budgets $\varepsilon = \{5,10,25\}$ to demonstrate the relevance of directing substitution to words that share similar syntactic roles rather than restricting substitution only to words that appear in a similar semantic context. 

These privacy budgets represent three privacy regimes: $\varepsilon=5$ for high privacy, $\varepsilon=10$ for moderate privacy, and $\varepsilon=25$ for low privacy. 



\subsection{Linguistic Analysis}
\label{sec:linguistics}

We intend to assess the effectiveness of our constraint to the candidate selection in retaining grammatical properties of words after substitution. We query each word contained in the vocabulary $100$ times and record the grammatical category for its surrogate word in the form of a frequency count. 

\begin{figure}[!t]
    \centering
    \subfigure[\texttt{MADLIB} with $k=1$]
    {
        \includegraphics[width=0.4\textwidth]{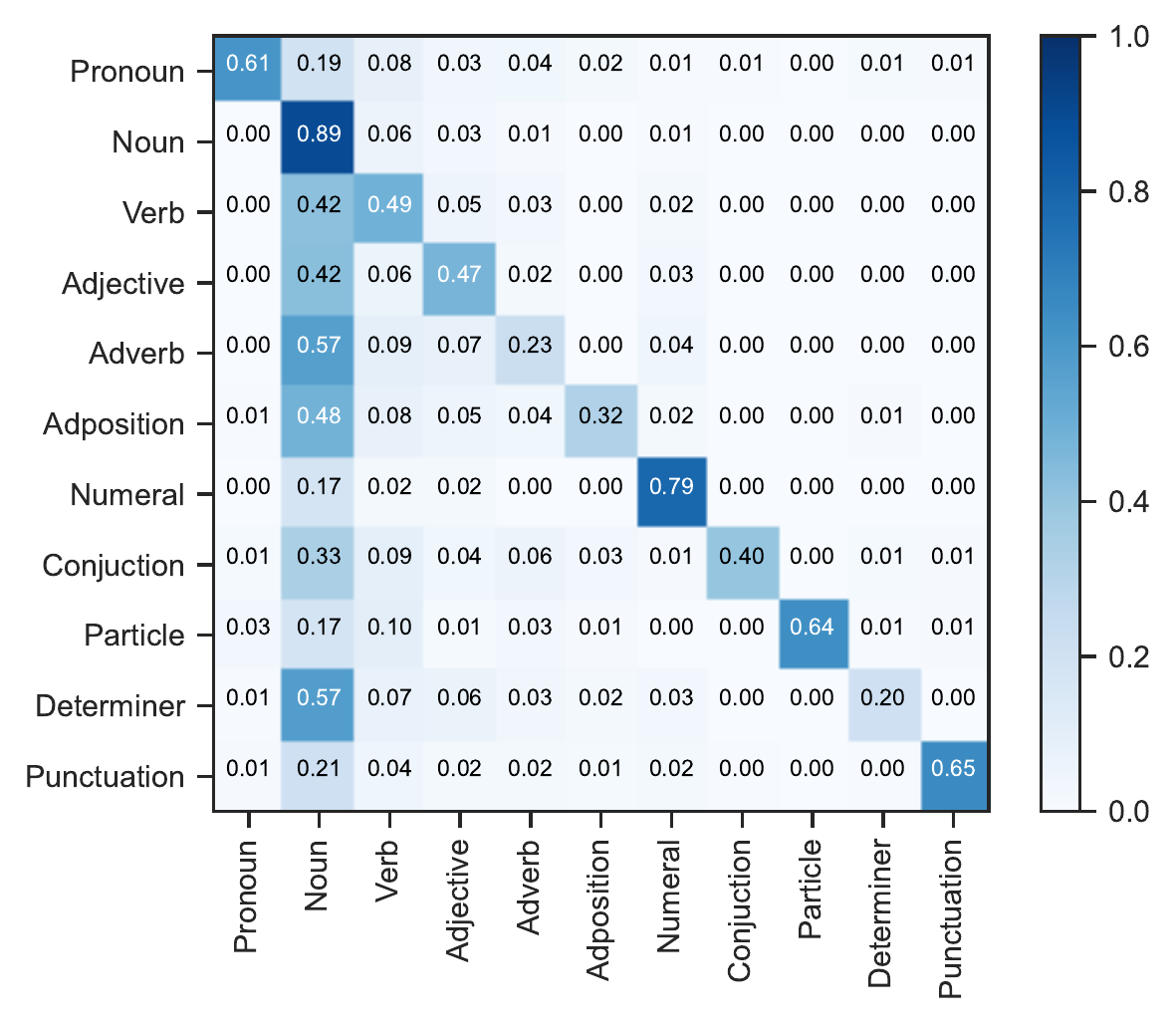}
        \label{fig:lg}
    }
    \hfill
    \subfigure[\texttt{MADLIB} with $k=20$]
    {
        \includegraphics[width=0.4\textwidth]{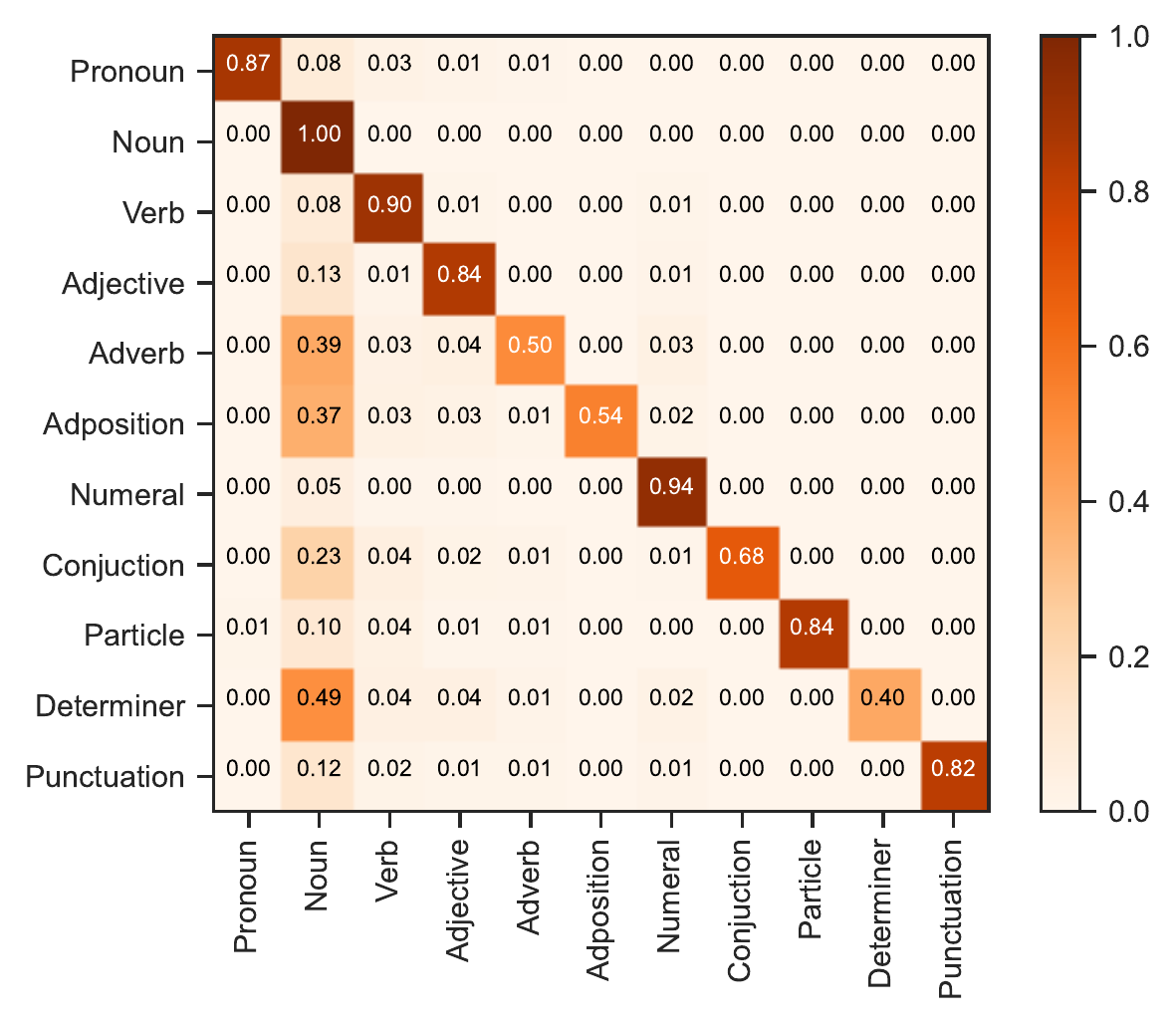}
        \label{fig:ll}
    }
    \caption{Approximated frequency counts by querying a subset of words and recording their universal part-of-speech tags before and after substitution. The diagonal represents the ideal preservation of grammatical properties.}
    \label{fig:linguistics}
\end{figure}

Given a moderate privacy budget of $\varepsilon = 10$, Figure \ref{fig:linguistics} visualizes the calculated frequency counts similar to a confusion matrix. The diagonal represents the preservation capability of grammatical categories, \ie, universal part-of-speech tags. A comparison across $\varepsilon \in \{5,10,25\}$ is deferred to Figure \ref{fig:linguistics_appendix} in the Appendix \ref{appendix:analysis}. 

We start with the examination of the baseline mechanism in Figure \ref{fig:lg}. Consistent with the independent and concurrent results of \citet{mattern2022limits}, our results indicate that the privatization mechanism is likely to cause grammatical errors. \citet{mattern2022limits} estimate that the grammatical category changes in $7.8\%$, whereas we calculated about $45.1\%$ for an identical privacy budget. This difference arises from the fact that \citet{mattern2022limits} only consider the four most frequent categories of \textit{nouns}, \textit{verbs}, \textit{adjectives}, and \textit{adverbs}, while we consider eleven categories according to the universal part-of-speech tagset. In addition to the number of grammatical categories, we indicate the fluctuations between categories, while \citet{mattern2022limits} only measures whether a category was changed. Owing to the tracking of the fluctuations, we find a disparate impact on the preservation of the grammatical categories. We find that the preservation of grammatical categories of words declines with growing guarantees for privacy, until the text after privatization consist almost entirely of nouns.

We compare these results to our constrained mechanism in Figure \ref{fig:ll}. With the introduction of a constrained candidate pool of size $k=20$, we observe an increased likelihood that surrogate texts retain the grammatical structure of the original texts. This can be seen by the dominance of the vertical line in Figure \ref{fig:lg} compared to initial signs of a diagonal line in Figure \ref{fig:ll}. Compared to the baseline value $45.1\%$, the preservation capability bounds at $81.4\%$.

\begin{figure}[!t]
    \centering
    \subfigure[\texttt{MADLIB} with $k=1$]
    {
        \includegraphics[width=0.33\textwidth]{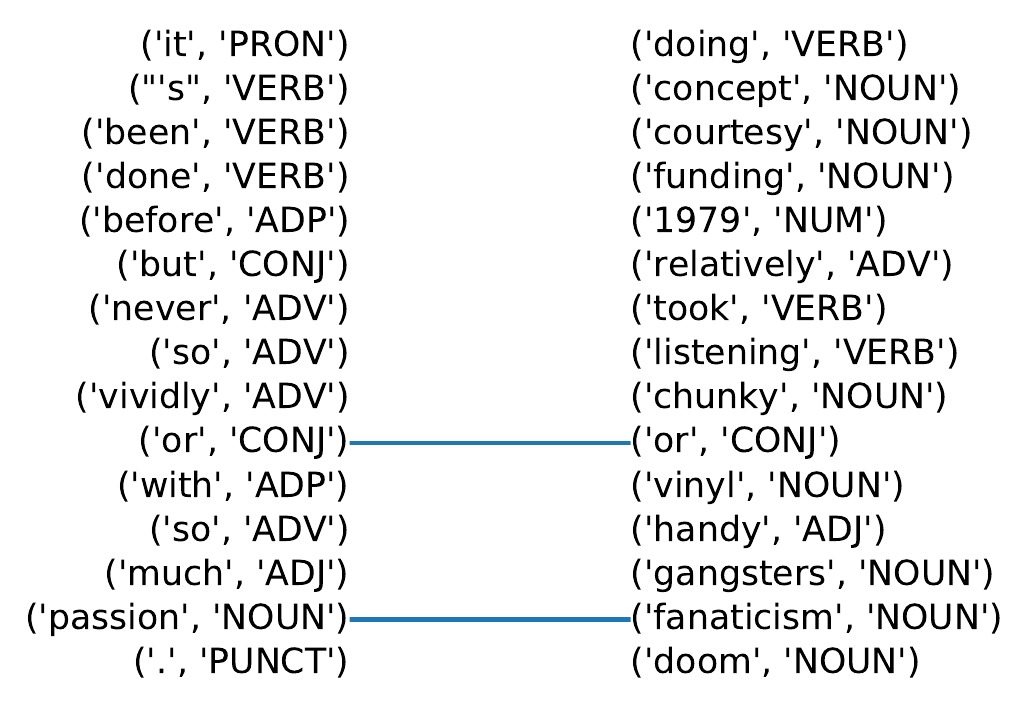}
        \label{fig:lg}
    }
    \hfill
    \subfigure[\texttt{MADLIB} with $k=20$]
    {
        \includegraphics[width=0.33\textwidth]{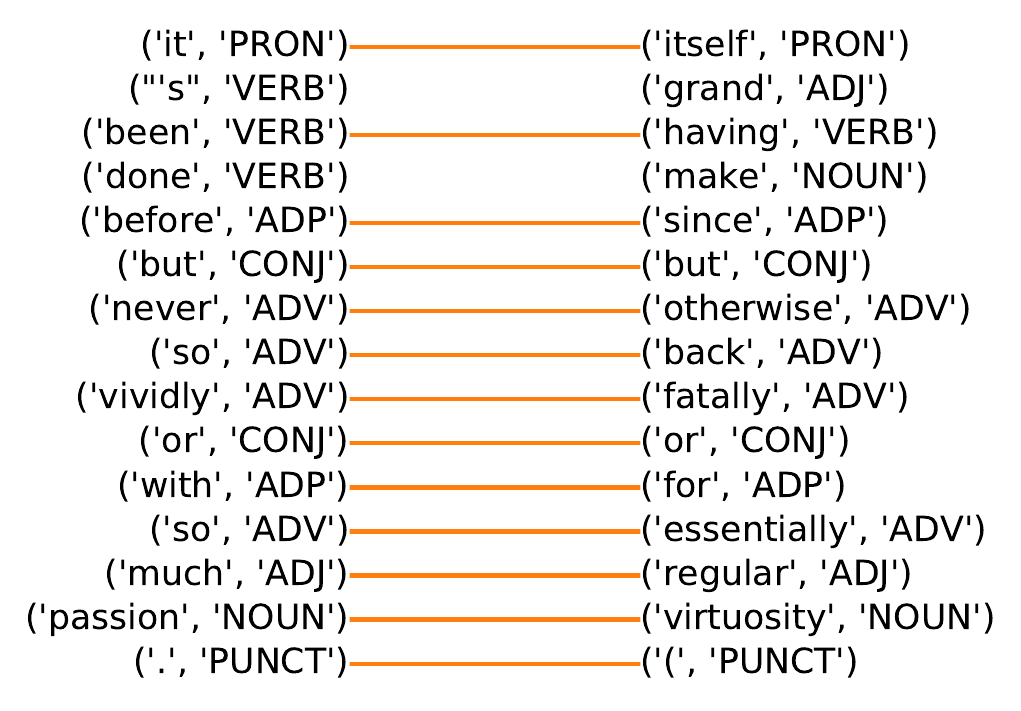}
        \label{fig:sl}
    }
    \caption{Example of syntax-preserving capabilities of \texttt{MADLIB} with and without grammatical constraint.}
    \label{fig:example}
\end{figure}

We illustrate the alignment of grammatical properties between words from a sensitive text and and their surrogate words with an example sentence in Figure \ref{fig:example}. We note that our syntactic guidance prevents words from being misleadingly replaced by numbers (and vice versa), as in the case of \textit{before} being replaced by \textit{1979}.

\subsection{Geometric Analysis} \label{sec:geometrics}

Intuitive properties for analyzing a mechanism operating on embeddings include magnitude, direction, and orthogonality. Since embeddings capture word co-occurrences, we expect most substitutions to be located in the same region of an embedding space and in the same direction from the embedding origin. 

We aim to measure the differences in the Euclidean distance of words with those of their corresponding substitutes generated by baseline $\mathcal{M}(w)$ and our constraint $\mathcal{M}^{'}(w)$. The results capture $\| w - \hat{w} \|$ and $\| w - \hat{w}^{'} \|$, respectively. Since the distances are zero when $w = \hat{w}$ or identical when $\hat{w} = \hat{w}^{'}$, we are only interested in the distances when a substitution has occurred and the mechanisms decided  on a distinct candidate for their substitution, \ie, $\mathcal{M}(w) \neq \mathcal{M}^{'}(w) \neq w$.

\begin{figure}[!t]
    \includegraphics[width=0.48\textwidth]{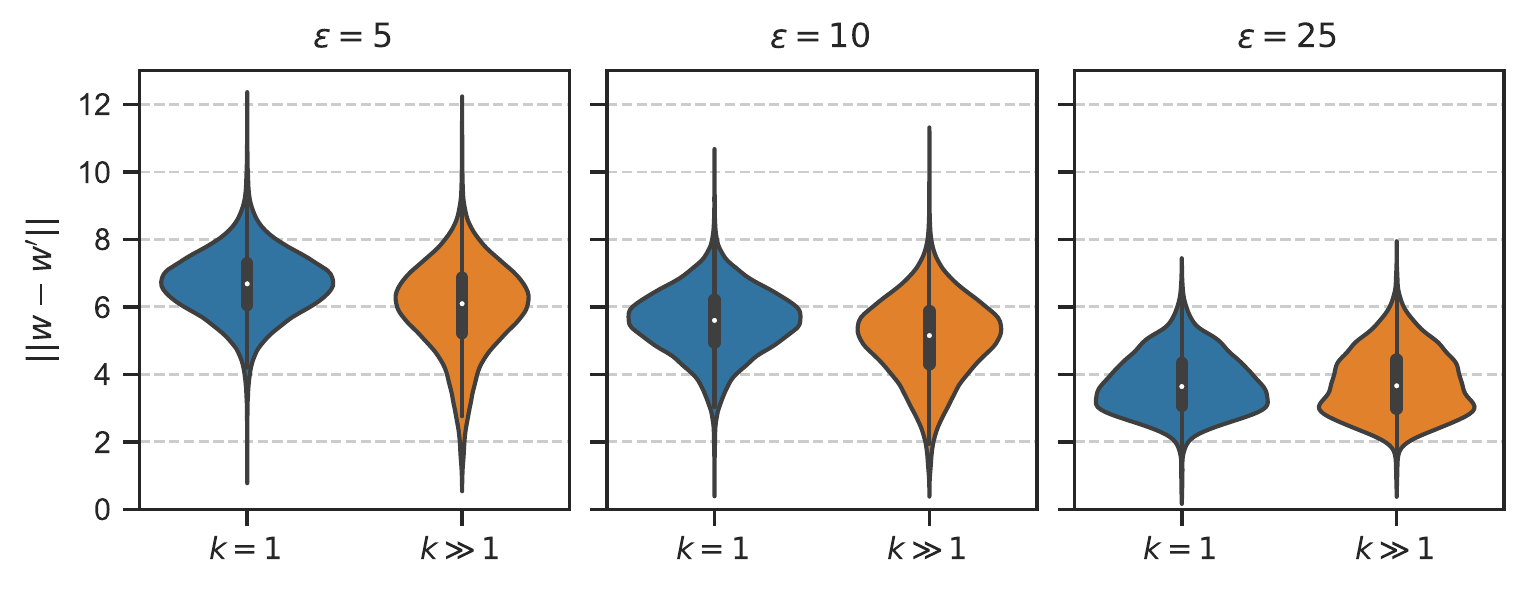}
    \caption{Euclidean distance for word substitutions. We depict default \texttt{MADLIB} ($k=1$) in blue and \texttt{MADLIB} ($k=20$) with grammatical constraint in orange.}
    \label{fig:distance}
\end{figure}

Figure \ref{fig:distance} depicts the calculated distances for querying words from our subset $100$ times. The distance approximation was carried out at a strategically chosen discrete set of values of $\varepsilon = \{5,10,25\}$. Since the distance is calculated as the difference between words and their substitutes, lower values indicate better substitutions. The distances depend on the amount of noise injected into the randomized mechanisms. The more noise, the larger the distances. Apparent across all privacy budgets, the distances between words and their substitutions are slightly shifted towards smaller distances. Since the distributions of distances are almost identical, we can take a principled guess that substitution in both mechanisms generally occurs within a similar region of the embedding space.

\subsection{Privacy Analysis} \label{sec:privacy}

Confronted with a non-zero probability that the candidate pool contains the sensitive word and no other word exists in the candidate pool with matching grammatical properties, it could be argued that the privacy guarantees suffer from the increased risk of self-substitution. By calculating the plausible deniability \citep{bindschaedler2017plausible}, we evaluate the risk of self-substitution arising from our grammatically constrained candidate selection.

In line with previous studies on text-to-text privatization \citep{feyisetan2019leveraging, feyisetan2020privacy, xu2021utilitarian, qu2021privacy}, we record the following statistics as proxies for plausible deniability.

\begin{enumerate}

\item[$\bullet$] $N_w = \mathbb{P} \{ M(w) = w \}$ measures the probability that a word is not substituted by the mechanism. This is approximated by counting the number of times a word $w$ is substituted by the same word after running the mechanism $100$ times.

\item[$\bullet$] $S_w = |\mathbb{P} \{ M(w) = w^{'} \}|$ measures the effective support in terms of the number of distinct substitutions produced for a word from the mechanism. This is approximated by the cardinality of the set of words $w^{‘}$ after running the mechanism $100$ times.

\end{enumerate}

Since the noise is scaled by $\nicefrac{1}{\varepsilon}$, we can make a connection between the proxy statistics and the privacy budget $\varepsilon$. A smaller $\varepsilon$ corresponds to a more stringent privacy guarantee. Adding more noise to the vector representation of a word results in fewer self-substituted words (lower $N_w$) and a more diverse set of distinct substitutions (higher $S_w$). A higher $\varepsilon$ corresponds to a less stringent privacy guarantee. This translates into less substitutions (higher $N_w$) and a narrow set of distinct substitutions (lower $S_w$). From a distributional perspective, it follows that $N_w$ ($S_w$ should be positively (negatively) skewed to provide reasonable privacy guarantees.

\begin{figure}[!t]
    \subfigure[$N_w$ refers to the number of substitute words that are \textit{identical} to a queried sensitive word.]
    {
        \includegraphics[width=0.48\textwidth]{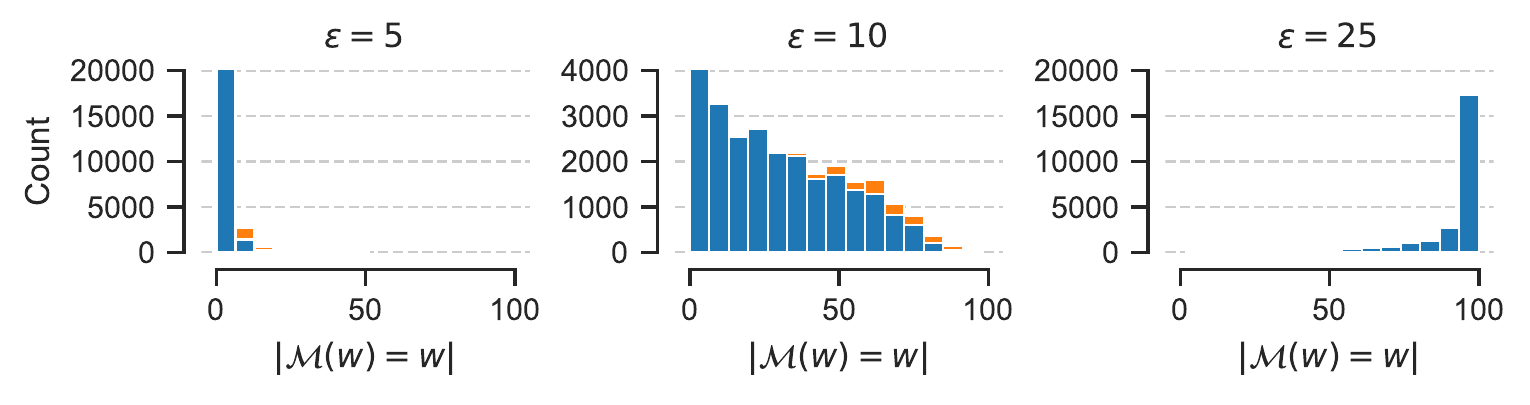}
        \label{fig:nw}
    }
    \hfill
    \subfigure[$S_w$ refers to the number of substitute words that are \textit{unique} from a queried sensitive word.]
    {
        \includegraphics[width=0.48\textwidth]{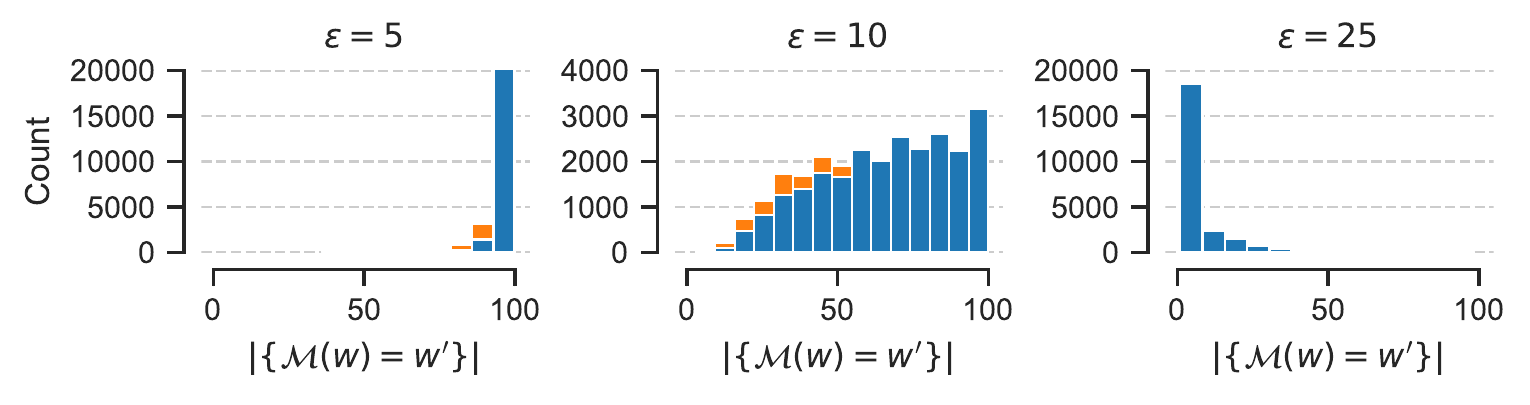}
        \label{fig:sw}
    }
    \caption{
        Plausible deniability statistics approximated for a carefully compiled sub-vocabulary of $24,525$ words of varying lexical categories, with each word independently privatized over a total number of $100$ queries. We present the baseline in blue and highlight the distribution shift induced by the grammatical constraint in orange.
    }
    \label{fig:privacy}
\end{figure}

For privacy budgets of $\varepsilon = \{5,10,25\}$, we present the distribution of $N_w$ and $S_w$ over $100$ independent queries Figure \ref{fig:privacy}. While lower values of $\varepsilon$ are desirable from a privacy perspective, it is widely known that text-to-text privatization requires slightly larger privacy budgets to provide reasonable utility in practice. Values of $\varepsilon$ up to 20 and 30 have been reported in related mechanisms \citep{feyisetan2020privacy}. The histograms serve as visual guidance for comparing (and selecting) the required privacy budget $\varepsilon$. As both mechanisms build upon the Euclidean distance as a metric, their privacy guarantees should match by using the same privacy budget $\varepsilon$. Directing the substitution to words with a matching grammatical category result in marginal changes to the plausible deniability. This is visually recognizable by the distribution shift. The grammatical constraint risks slightly more self-substitutions and reduced effective support. This is because words are substituted (almost) only by words from the same grammatical category, reducing the pool of unique words that are appropriate for substitution and thus reducing the effective support of the multivariate mechanism. Out of $100$ words queried given a fixed privacy budget of $\varepsilon=10$, self-substitution increases on average from about $29$ to $32$, while effective support decreases on average from about $66$ to $61$. The fact that both changes in $N_w$ and $S_w$ do not exceed or fall below $50$ indicates that plausible deniability is assured for the average-case scenario. We conclude that the grammatically constrained candidate selection does not come at the expense of privacy and can therefore be incorporated into the privatization step without the need to recalibrate the proxies for plausible deniability.

\begin{table*} 
\centering
\resizebox{\linewidth}{!}{%
\begin{tabular}{ccllllllllllll} 
\toprule
                                 &                                                                                              & \multicolumn{2}{c}{\textbf{Classification}}           & \multicolumn{1}{c}{} & \multicolumn{3}{c}{\textbf{Textual Similarity}}                                   & \multicolumn{1}{c}{} & \multicolumn{3}{c}{\textbf{Textual Entailment}}                                   & \multicolumn{1}{c}{} & \multicolumn{1}{c}{\textbf{Avg}.}       \\ 
\cline{3-4}\cline{6-8}\cline{10-12}\cline{14-14}
                                 & \multirow{2}{*}{\begin{tabular}[c]{@{}c@{}}\textbf{Level of}\\\textbf{Privacy}\end{tabular}} & \multicolumn{1}{c}{CoLA}  & \multicolumn{1}{c}{SST2}  & \multicolumn{1}{c}{} & \multicolumn{1}{c}{QQP}   & \multicolumn{1}{c}{MRPC}  & \multicolumn{1}{c}{STSB}  & \multicolumn{1}{c}{} & \multicolumn{1}{c}{MNLI}  & \multicolumn{1}{c}{QNLI}  & \multicolumn{1}{c}{RTE}   & \multicolumn{1}{c}{} & \multicolumn{1}{c}{\multirow{2}{*}{-}}  \\
                                 &                                                                                              & \multicolumn{1}{c}{(MCC)} & \multicolumn{1}{c}{(ACC)} & \multicolumn{1}{c}{} & \multicolumn{1}{c}{(ACC)} & \multicolumn{1}{c}{(ACC)} & \multicolumn{1}{c}{(SCC)} & \multicolumn{1}{c}{} & \multicolumn{1}{c}{(ACC)} & \multicolumn{1}{c}{(ACC)} & \multicolumn{1}{c}{(ACC)} & \multicolumn{1}{c}{} & \multicolumn{1}{c}{}                    \\ 
\midrule
\textbf{BERT} & - & 0.5792  & 0.9243  & & 0.8879 & 0.8329 & 0.8854 & & 0.8229  & 0.8912 & 0.6927 & & 0.8146 \\ 
\hline

\multirow{2}{*}{$k=1$} 
& \multicolumn{1}{l}{$p=0.9$} & 0.0248  & 0.8127  & & 0.6940 & 0.5603 & 0.6153 & & 0.5304  & 0.6327 & \textbf{0.5663} & & 0.5545 \\
& \multicolumn{1}{l}{$p=0.5$} & 0.2303  & 0.8848  & & 0.8181 & 0.6242 & 0.7951 & & 0.7114  & 0.8339 & 0.6027 & & 0.6875 \\

\multirow{2}{*}{$k=20$} 
& \multicolumn{1}{l}{$p=0.9$} & \textbf{0.0928}  & \textbf{0.8510}  & & \textbf{0.7519} & \textbf{0.5946} & \textbf{0.6988} & & \textbf{0.6251}  & \textbf{0.7423} & 0.4525 & & \textbf{0.6011} \\
& \multicolumn{1}{l}{$p=0.5$} & 0.3493  & 0.9035  & & 0.8397 & 0.6333 & 0.8011 & & 0.7301  & 0.8627 & 0.5420 & & 0.7077 \\ 
\bottomrule
\end{tabular}
}
\caption{Results on a subset of \texttt{GLUE} \citep{wang2019glue}.  We report Matthews correlation for the \texttt{CoLA} dataset, Spearman correlation for the \texttt{STSB} dataset, and the accuracy score for all remaining datasets. The level of privacy increases with the quantile of words that are provable plausible deniable. $p=.90$ denotes an (almost) worst-case scenario. $p=.50$ denotes an average-case scenario. We fixed the candidate pool to $k=20$. A candidate pool of $k=1$ reduces to the randomized mechanism of \citet{feyisetan2020privacy}. Bold font indicates the best result from three independent trials of the worst-case scenario.}
\label{tab:benchmark}
\end{table*}

Rather than compromising privacy, our constrained candidate selection can be alternatively viewed as a barrier against reconstruction attacks. Recall that the nearest neighbor search is generalized from $k=1$ to $k\gg1$. This generalization may impede naïve inversion attacks such as the one proposed in \citet{song2020information}, in which an adversary attempts to recover a word by finding the nearest neighbor to the substitute word. Although this inversion attack is not comprehensive, it can be used as a reference point for investigations regarding the robustness of privacy attacks. We include the setup and the results of a membership inference attack in the Appendix \ref{appendix:privacy}.

\subsection{Utility Analysis} \label{sec:utility}

To evaluate whether the preservation of syntactic roles translates to better utility in downstream tasks, we conduct experiments with \texttt{BERT} \citep{devlin2018bert} on a subset of \texttt{GLUE} \citep{wang2019glue}. 




Once for each mechanism under comparison, we privatize the training corpus of each dataset. Since the privacy guarantees do not exactly match, we calculate the available privacy budget for each mechanism such that the $.90$ quantile of words is plausible deniable. This resembles a practical scenario where we allow a negligible subset of words without provable privacy guarantees. 


We report the performance scores in Table \ref{tab:benchmark}. A baseline trained on unprotected data is listed as an upper bound on the performance. All trials mimic the training of the baseline. To privatize the texts in the datasets, we use our modification with a varying candidate pool of size $k \in {1,20}$. Recall that $k=1$ reduces our modification to the multivariate mechanisms of \citet{feyisetan2020privacy}. Although we focus our analysis on a worst-case scenario in which the $.90$-quantile of words is plausibly deniable, we included test results for an average-case scenario in which only a $.50$-quantile of words enjoys plausible deniability.

On average, \texttt{BERT} bounds at $81.46\%$ when trained on sensitive text. Compared to the baseline, \texttt{BERT} trained on surrogate texts attains $55.45\%$ when the candidate pool is $k=1$. By broadening the candidate pool to $k=20$ and directing the substitution to words with matching grammatical categories, \texttt{BERT} trained on surrogate texts ranks at $60.11\%$. This corresponds to narrowing down the performance loss by $4.66\%$. 

Contrary to our initial assumption that preserving the syntactic role of words is particularly relevant to sentiment analysis, we find evidence that accounting for syntactic information during privatization benefits a variety of downstream tasks. We conclude that linguistic guidance is a legitimate alternative perspective to previous extensions that focus on the geometric position of words in the embedding.

\section{Conclusion}
\label{section:5}

Privatizing written text is typically achieved through text-to-text privatization over the embedding space. Since text-to-text privatization scales the notion of indistinguishably of differential privacy by a distance in the geometric space of embeddings, prior studies focused on geometric properties \citep{feyisetan2019leveraging, xu2020differentially, carvalho2021tem}.

Unlike prior studies on amplifying text-to-text privatization by accounting for the geometric position of words within the embedding space, we initialized a set of strategies for amplification from the perspective of grammatical properties, such as \textit{category}, \textit{number}, or \textit{tense}.

By incorporating grammatical properties in the form of part-of-speech tags into text-to-text privatization, we direct the privatization step towards preserving the syntactic role of a word in a text. We experimentally demonstrated that that surrogate texts that conform to the structure of the sensitive text outperform surrogate texts that strictly rely on co-occurrences of words in the embedding space. 



\paragraph{Limitations.} We note that directing the substitution to candidates with matching grammatical categories incurs additional information leakage that is not accounted for by our modification. Too remedy the unaccounted information leakage, one could recast the candidate selection through the exponential mechanism \citep{mcsherry2007mechanism}.

\section*{Acknowledgment}

We gratefully acknowledge that this research was supported in part by the \textit{German Federal Ministry of Education and Research} through the \textit{Software Campus} (ref. \textit{01IS17045}).

\bibliography{submission}
\bibliographystyle{acl_natbib}



\appendix

\clearpage
\onecolumn

\section*{Appendices}
\label{sec:appendix}

\renewcommand{\thesubsection}{\Alph{subsection}}
\renewcommand{\thefigure}{\Alph{subsection}.\arabic{figure}}


\setcounter{figure}{0}
\setcounter{table}{0}

\subsection{Linguistic Evaluation} \label{appendix:analysis}

\begin{figure*}[ht]
    \centering
    \subfigure[\texttt{MADLIB} with $k=1$]
    {
        \includegraphics[width=0.8\textwidth]{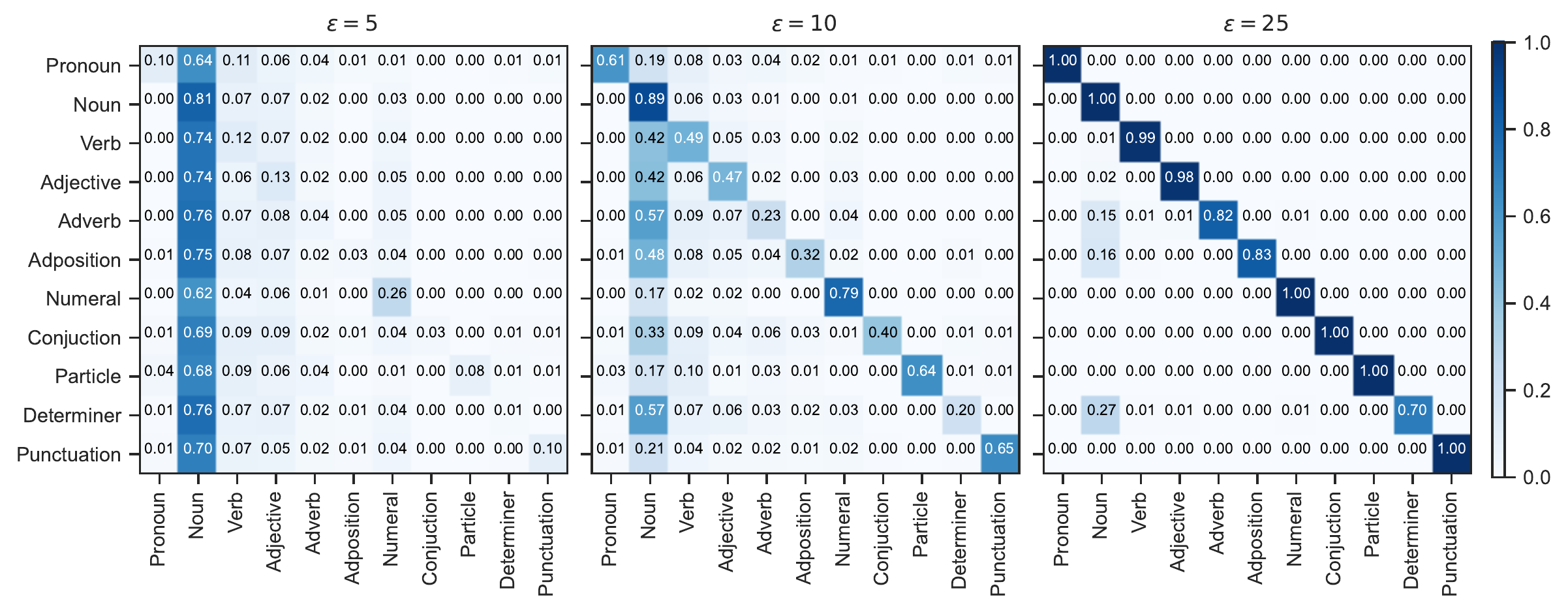}
        \label{fig:lg_appendix}
    }
    \hfill
    \subfigure[\texttt{MADLIB} with $k=20$]
    {
        \includegraphics[width=0.8\textwidth]{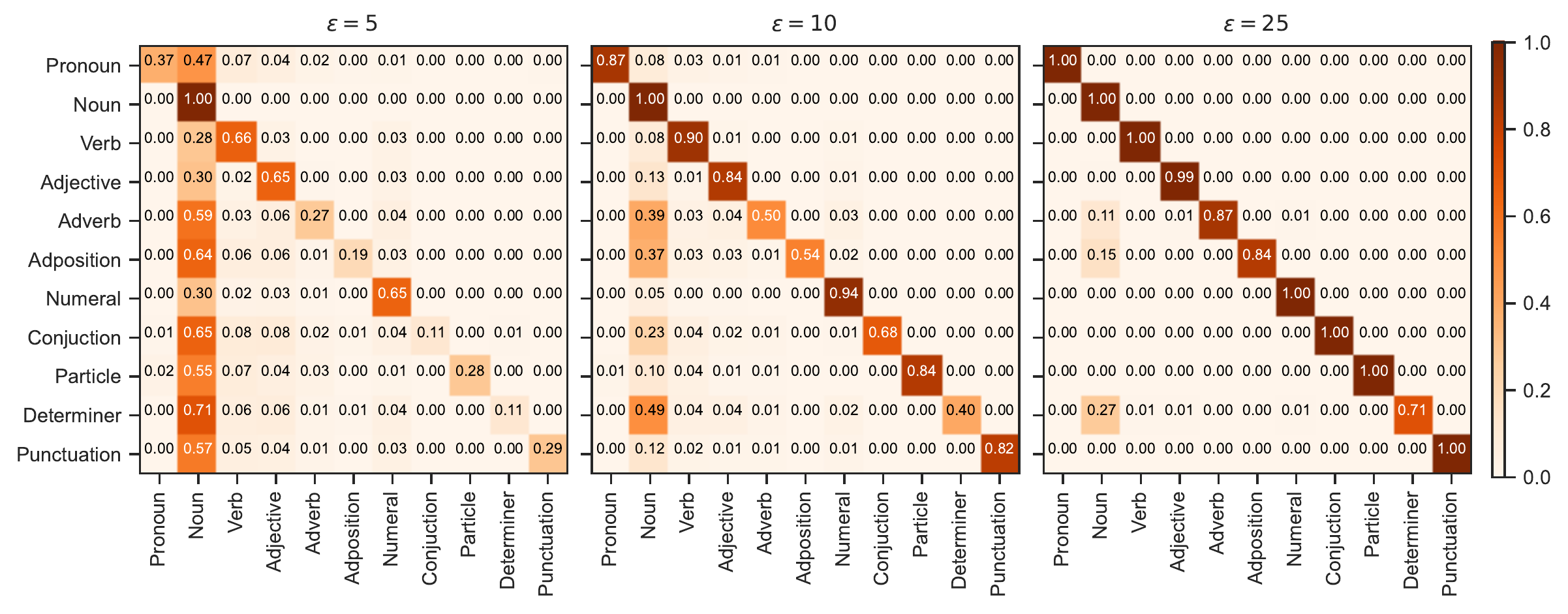}
        \label{fig:ll_appendix}
    }
    \caption{Linguistics analysis with respect to the grammatical category of a sub-vocabulary after $100$ times of querying a randomized mechanism. Given a candidate pool $k$ of nearest neighbors, $k=1$ represents substitutions solely based on co-occurrences, whereas $k=20$ represents grammatically constraint substitutions. The size of the candidate pool has been approximated by the sub-vocabulary's neighborhood.}
    \label{fig:linguistics_appendix}
\end{figure*}

Covering three levels of privacy budgets $\varepsilon$, we include the detailed linguistics analysis of the multivariate substitutions obtained from \texttt{MADLIB} \citep{feyisetan2020privacy} in Figure \ref{fig:linguistics_appendix}. 

Without a constraint on syntactic roles, we cannot expect the privatization step to yield surrogate texts that conform to the structure of the sensitive texts. From the diagonal, it can be clearly seen that our grammatical constraint retains most grammatical categories across all budget budgets and all types of categories. At a low privacy budget of $\varepsilon=5$, the preservation capability of grammatical categories is $0.4163$. At a moderate privacy budget of $\varepsilon = 10$, the preservation capability bounds at $0.8145$. At a high privacy budget of $\varepsilon = 25$, the advantage in the preservation capability diminishes as the perturbation probability in general decreases. 

\subsection{Setup and Results from Membership Inference Attack} \label{appendix:privacy}

To eliminate the possibility that the performance gain is caused by mismatching privacy guarantees, we perform a \textit{Membership Inference Attack} (MIA) introduced by \citet{shokri2017membership}. Given black-box access to a model, an adversary attempts to infer the presence of records from an inaccessible training corpus. We follow the experimental setup of \citet{carvalho2021tem} for our membership inference attack. To maximize the attack uncertainty, we divide the \texttt{IMDb} dataset into four \textit{disjoint} partitions with an equal number of members and non-members, respectively. The target model is trained on the first partition after privatization by each mechanism, whereas the shadow model is trained on the non-privatized second partition. The shadow model architecturally mimics the target model. We then build an attack model composed of a two-layer multi-layer perception with a hidden size of $64$ and non-linear activations. To train the attack model, we feed the logits obtained by the second and third partitions given by the shadow model, where logits from the second first partition are labeled as members and logits from the third partition are labeled as non-members. Once the attack model is trained, we feed the logits of the first partition and the fourth partition obtained by the target model, where logits from the first partition are labeled as members and logits from the fourth partition are labeled as non-members.

We measure the success rate of our membership attack using macro-averaged metrics for precision and recall. Precision captures the fraction of records for which the membership was correctly inferred. Recall captures the coverage of the membership attack. Since the baseline accuracy of the membership attack is $0.5$, we consider a randomized mechanism to be provably private if and only if it holds the attack accuracy close to that of random guessing. We report the attack accuracy as the area under the precision-recall curve. We report a non-private membership accuracy of $0.53$. Given a practical privacy budget, both mechanisms fluctuate around the $0.5$ mark averaged across three independent trials. With no clear hint, we thus conclude that the performance gain induced by a grammatical constraint cannot be attributed to a latent privacy loss.

\end{document}